\documentclass{article}
\usepackage{graphicx}
\usepackage{subfigure}
\usepackage{here}
\usepackage[margin=1.25in]{geometry}
\usepackage{authblk}
\usepackage{url}

\usepackage{listings}
\usepackage{xcolor}

\definecolor{identifier}{rgb}{0.611, 0.862, 0.996}  
\definecolor{comment}{rgb}{0.415, 0.600, 0.333}     
\definecolor{keyword1}{rgb}{0.768, 0.521, 0.749}    
\definecolor{keyword2}{rgb}{0.862, 0.862, 0.666}    
\definecolor{keyword3}{rgb}{0.2, 0.8, 1}            
\definecolor{keyword4}{rgb}{0.854, 0.439, 0.839}    
\definecolor{string}{rgb}{0.847, 0.560, 0.458}      

\begin{document}

\title{Inductive-bias Learning : Generating Code Models with Large Language Model}

\author[1*]{Toma Tanaka}
\author[1‡]{Naofumi Emoto}
\author[1††]{Tsukasa Yumibayashi}

\affil[1]{BrainPad inc., 3-1-1 Roppongi, Minato-ku, Tokyo 108-0071, Japan.}
\affil[*]{toma.tanaka@brainpad.co.jp}
\affil[$\dag$]{naofumi.emoto@brainpad.co.jp}
\affil[$\ddag$]{tsukasa.yumibayashi@brainpad.co.jp}

\date{\empty}

\maketitle

\begin{abstract}
Large Language Models(LLMs) have been attracting attention due to a ability called in-context learning(ICL). 
ICL, without updating the parameters of a LLM, it is possible to achieve highly accurate inference based on rules ``in the context'' by merely inputting a training data into the prompt. 
Although ICL is a developing field with many unanswered questions, LLMs themselves serves as a inference model, seemingly realizing inference without explicitly indicate ``inductive bias''. 
On the other hand, a code generation is also a highlighted application of LLMs. 
The accuracy of code generation has dramatically improved, enabling even non-engineers to generate code to perform the desired tasks by crafting appropriate prompts. 
In this paper, we propose a novel ``learning'' method called an ``Inductive-Bias Learning (IBL)'', which combines the techniques of ICL and code generation.
An idea of IBL is straightforward. 
Like ICL,  IBL inputs a training data into the prompt and outputs a code with a necessary structure for inference (we referred to as  ``Code Model'') from a ``contextual understanding''. 
Despite being a seemingly simple approach, IBL encompasses both a ``property of inference without explicit inductive bias'' inherent in ICL and a ``readability and explainability'' of the code generation.
Surprisingly, generated Code Models have been found to achieve predictive accuracy comparable to, and in some cases surpassing, ICL and representative machine learning models. Our IBL code is open source:
\url{https://github.com/fuyu-quant/IBLM}

\end{abstract}

\section{Inroduction}
\begin{figure}[H]
  \centering
  \includegraphics[width=0.9\textwidth]{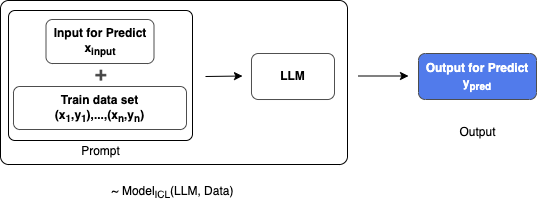}
  \caption{In-Context Learning}
  \label{fig:icl}
\end{figure}

Scaled up Language models ``understand'' data relationships ``in the context'' by inputting some a training data(e.g., input-output pairs) directly to a prompt without fine-tuning. 
Moreover, it has been shown to perform inference with high accuracy and it is expected applicable to a wide range of tasks by characteristics of LLMs \cite{brown2020language}. 
This is a property of a large language models(LLM) called In-Context Learning(ICL)(eq.\ref{eq:icl}, Figure \ref{fig:icl}):

\begin{equation}
\mathrm{LLM}.\mathrm{input {\_}prompt} \left [\mathcal{D}\left[ (x_1, y_1), \cdots, (x_n, y_n) \right], x_{\mathrm{input}} \right] \rightarrow y_{\mathrm{pred}} \label{eq:icl}
\end{equation}

\noindent Where, the left side of the figure is the prompt that expresses the input for the training data and the inference, and the right side of the figure is the output that expresses the inference result of the LLM.

ICL is a developing field, we give briefly review its development: First, as studies on its structures, (\cite{chan2022data}\cite{xie2021explanation}) describe necessary conditions of training data for LLMs to acquire ICL ability. 
Moreover, in \cite{akyurek2022learning}\cite{mahankali2023one}\cite{von2023transformers} a relationship between an inference process of LLMs by ICL and a gradient descent is studied. 
There are still many unclear points about structures of ICL, and further researches are required. 

Furthermore, as researches on its performance, \cite{garg2022can} shows that it is possible to learn various functions by ICL such as sparse linear functions, 2-layer neural networks, decision trees, etc. by pre-training Transformers. 
Therefore, an LLM that satisfies the appropriate conditions appears to ``understand'' an unknown data relationships by giving some examples of the inputs and the outputs according to an ability of ICL and can predict the value of the output value for the input.

Now we encounter a natural question related to ICL:

\bigskip

\noindent ``Isn't it possible to directly the output the resulting ``model'' that LLMs learned from the input and the output examples?''

\bigskip

\begin{figure}[H]
  \centering
  \includegraphics[width=1\textwidth]{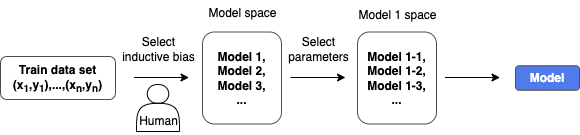}
  \caption{Traditional Machine Learning  
  with inductive bias by Human Selection}
  \label{fig:ib}
\end{figure}

If the above issues can be resolved, we can explicitly show what data relationships LLM has found from the input and output examples.
Furthermore, we can eliminate  the ``traditional machine learning with inductive bias by human selection''(Figure \ref{fig:ib}) and make direct inferences like ICL.

This also means that LLM can be thought of as ``learning how to learn'' from a data, and can handle a perspective one step higher than ICL. 
This can be interpreted as bringing a ``Meta-learning'' perspective to ICL.

To solve this question, we formulate the problem concretely as follows:

\bigskip

\noindent ``Can LLM take a training data (here, a pair of explanatory and objective variables) as input and output a model for inferring the objective variable from the explanatory variables?''

\bigskip

\begin{figure}[H]
  \centering
  \includegraphics[width=1\textwidth]{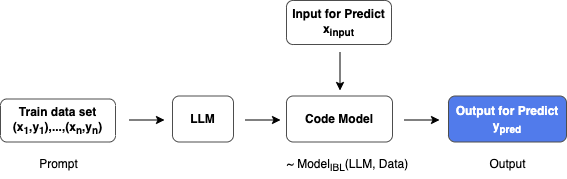}
  \caption{Inductive-Bias Learning}
  \label{fig:ibl}
\end{figure}

Our answer to this proposition is ``YES''. We named this method an Inductive-bias Learning (IBL) (eq. \ref{eq:ibl}, Figure \ref{fig:ibl}). The name the ``Inductive-bias'' Learning comes from outputting models with various structures for each data without explicitly assuming the inductive bias.

\begin{equation}
\mathrm{LLM}.\mathrm{input {\_}prompt} \left [\mathcal{D}\left[ (x_1, y_1), \cdots, (x_n, y_n) \right] \right] \rightarrow \mathrm{Code \ Model} _{\mathrm{LLM}, \mathcal{D}}(x_{\mathrm{input}}) \rightarrow y_{\mathrm{pred}}
\label{eq:ibl}
\end{equation}

In this paper, we performed the following verification: We set binary classification problems as a task,  and the GPT-4 as an LLM. Moreover we used multiple datasets and seed values to sample the data, using IBL and representative machine learning algorithms and a comparison of ICL accuracy were performed. As results, we find that IBL can provide accuracy comparable to or even better than typical machine learning algorithms and ICL in some cases.

As given in the proposition, IBL outputs the inference structure itself from the training data. 
Moreover, this can be done for a variety of data. 
This suggests the possibility that LLM learns how to learn the relationship between input and output when given data, and LLM also has an aspect as a kind of Meta-Learning.

This paper is organized as follows: Section 2 summarizes related work on ICL and the Meta-Learning. Section 3 gives an architecture and training method of IBL using a LLM. Next, Section 4 presents two results of verifying the performance of IBL. One is accuracy verification with conventional major machine learning models, and the other is accuracy verification with ICL. Finally, Section 5, 6 summarizes the verification of IBL and the efforts and applicability required for future IBL.

\section{Related works}
There are two major areas of research related to IBL: ICL, and the Meta-Learning.

\subsubsection*{In-context Learning}
In the research by \cite{garg2022can}, ICL was demonstrated that by pre-training Transformer models, linear functions could be learned with accuracy comparable to the least squares estimator.
Furthermore, this approach has been shown to be applicable not only to linear functions but also to sparse linear functions, decision trees, and two-layer neural networks. 
In IBL of this study, high accuracy in prediction was achieved in several datasets, but the relationship with the learning ability of ICL, and whether IBL can learn linear functions, decision trees, and two-layer neural networks is unknown.

In \cite{wei2023larger}, the semantic prior distribution in ICL was removed, and accuracy testing was conducted using methods like Flipped-Label ICL (inverting the correct labels) or SUL-ICL (providing semantically unrelated labels). As a result, it became clear that if sufficient datas are provided, the semantic prior distribution can be overwritten. During the execution of IBL in this inverstigate, there is also a possibility that the Code Model is generated based on the provided training data by overwriting the semantic prior distribution if a sufficient number of data are given, but more detailed verification is needed.

\subsubsection*{Meta-Learning}
IBL might suggest that a certain type of the Meta Learning, represented by approaches like MAML \cite{finn2017model}, can be realized with LLMs. In IBL, structures for predicting unknown data are inferred without explicitly setting the inductive bias from the training data alone. This can be seen as LLMs being able to learn how to infer a logic for predicting from the data alone during its pre-training stage. This might, in a sense, suggest that LLMs are performing the Meta Learning within its learning process.

For instance, in the research of Meta ticket \cite{chijiwa2022meta}, it is possible to find a neural network structure that achieves high predictive accuracy in few-shot learning from data. In contrast, in this research, GPT-4 is used to find the structure of a model with high predictive accuracy only from data. While the aim of finding the optimal model structure is the same, the approaches and the data handled are different.

What happens internally when LLMs execute Inductive-bias Learning, and how LLM acquires the ability of IBL (Meta-Learning-like properties) during a learning process, is an intriguing theme. The inference by ICL being realized through gradient descent is suggested by researches such as \cite{garg2022can}, \cite{akyurek2022learning}, \cite{von2023transformers}, but how IBL is being realized is still not understood.

\section{Inductive-bias Learning}
\begin{figure}[ht]
  \centering
  \includegraphics[width=0.8\textwidth]{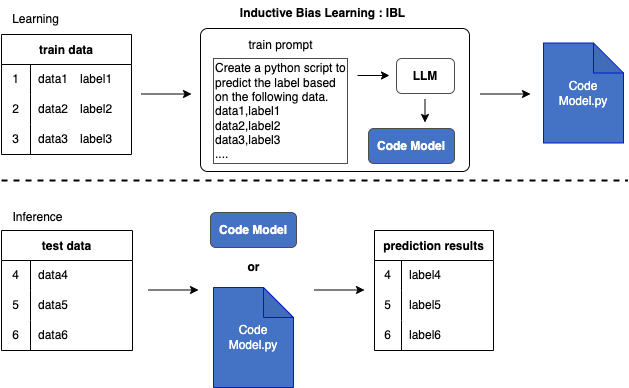}
  \caption{\\
  Above: Model generation by IBL: As in normal supervised learning, input training data consisting of multiple pairs of explanatory and objective variables. A code is created based on the input data. If necessary, a Python file is also created.\\
  Below: Prediction of a model generated by IBL: As with ordinary machine learning models, input the explanatory variables for the data you wish to predict. Make predictions using the generated code. Alternatively, you can execute the generated Python file to make predictions.}
  \label{fig:ibl_arch}
\end{figure}

The Inductive-bias learning (IBL) can be divided into two phases, learning and inference, similar to the usual supervised machine learning algorithms. 
The learning phase takes the input explanatory variable and objective variable pairs as inputs and determines the structures of the model for predicting the objective variable from the explanatory variables. 
In inference, the created model can be used to make predictions about unknown data. In this study, we focused on binary classification. 
Details are explained below(Figure \ref{fig:ibl_arch}).

\subsection{Learning}
A learning phase of IBL uses a given training data to generate a code what is represented as Python functions to output an objective variable from explanatory variables. If necessary, we also create a Python file in which the code is written. We call this generated Python code a ``Code Model''.

Moreover to a training data, which are input-output pairs, the prompts for an LLM include instructions to output a probability of label 1 for accuracy evaluation as a binary classification task, and to avoid including code that uses the general machine learning models in the generated Python code.
In this verification, the use of machine learning models was prohibited because we were interested in whether GPT-4 could output ``logic'' to make predictions from the data.

For an output, we generate a code for Pyhton function that takes the explanatory variable as input and the probability value what the label of the objective variable is 1 as output value. The generated Python code is generated as conditional branches by certain features or linear expressions involving multiple features. Various output results are also obtained depending on a dataset and a number of data used for a learning.

\subsection{Inference}
The inference phase of IBL is performed using the Python code generated by the Learning. The Python code that takes explanatory variables as inputs and probability values as  outputs is executed.

The Python code generated by IBL learning is very fast because it is generated for simple processes such as conditional branching by certain features or linear expressions involving multiple features. On the other hand, despite such simple processing, the generated Code Models have been verified to have very high prediction accuracy.

Furthermore, the Code Models generated by IBL are outputs as Python codes. This allows the user to see what features the model uses and how much they contribute to a forecast, making the model very easy to interpret.

\section{Experiments and Results}
In this section, we introduce two experiment results to test the performance of IBL. First, we compare the predictive ability of IBL and typical machine learning models on several datasets. Second, we compare the predictive ability of IBL and ICL on the ``Titanic dataset''.

\subsection{Experimental Setup}

\subsubsection*{Datasets and Task}
We compared accuracy of Code Models generated by IBL with well-known machine learning algorithms and ICL, focusing on the binary classification tasks. 
The datasets used in this evaluation are the well-known Titanic dataset from Kaggle, and pseudo and moon datasets available through scikit-learn.

For the Titanic dataset, we performed imputation using the mean for missing values and applied One-hot encoding for categorical data to align the inputs across all machine learning models. 
Additionally, we created training data by sampling with three different seed values to validate with several input datasets. 
Datas that was not used for training in each seed value was considered as test data. For the pseudo and moon data from scikit-learn, we generated synthetic data using three different seed values and split it into training and test data. 
Since having too many decimal places would reduce the number of data points that could be inputted into IBL or ICL, all generated values were rounded to three decimal places.
To eliminate the imbalance in the training data, we made the number of positive and negative examples the same.

\subsubsection*{Number of IBL executions}
One of the issues with IBL is the instability of outputs which are  generated Code Models; the Code Models generated by IBL do not always produce high prediction accuracy in a stable manner, and the accuracy varies. 
For this reason, we generated 30 Code Models for each dataset, training data size, and seed value, and compared the one with a highest AUC.

\subsubsection*{How to execute ICL}
To predict test data based on training data through ICL, the following method is utilized.
First, all the training datas are converted into a string and included in the prompt. 
Additionally, instructions are written into a prompt to output the explanatory variables of the data that you want to predict, along with the target variable for that data.
This prompt is input into GPT-4, and the target variable is produced as the output. 
These processes are carried out for all the test datas, and the resulting predictions are considered the output of ICL. 
The intention behind this method is to include as much training data as possible in the prompt.

In this paper, gpt-4-0613 was used for IBL and ICL. No additional training such as fine tuning was performed, and only prompt input was used.

\subsection{IBL vs Machine Learning Models}
\begin{figure}[H]
      \centering
      \subfigure[seed=3655]{
        \includegraphics[width=0.3\linewidth]{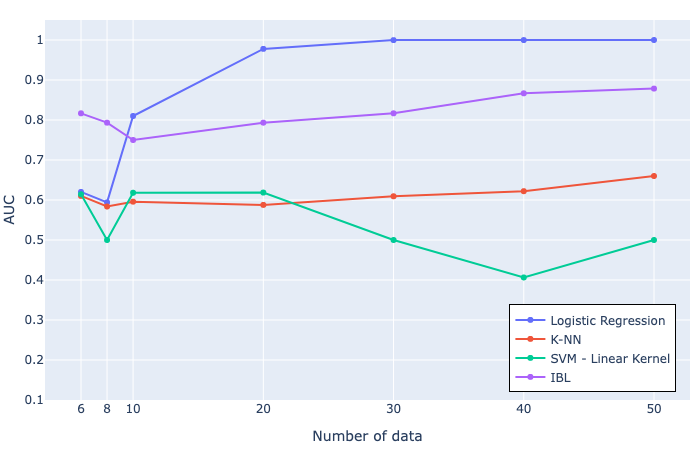}
        \label{fig:t1}
      }
      \subfigure[seed=3656]{
        \includegraphics[width=0.3\linewidth]{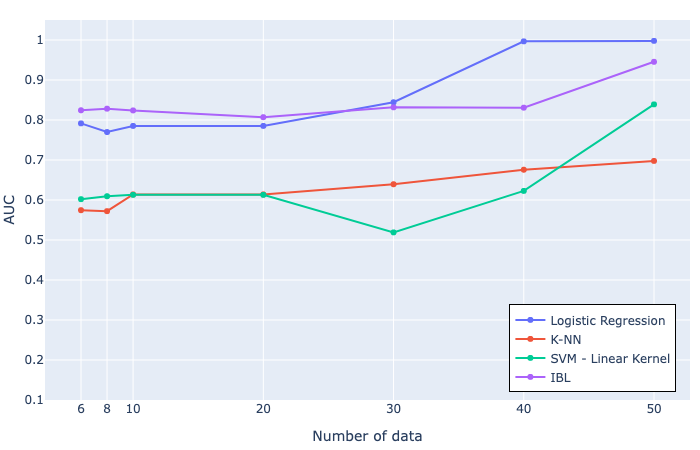}
        \label{fig:t2}
      }
      \subfigure[seed=3657]{
        \includegraphics[width=0.3\linewidth]{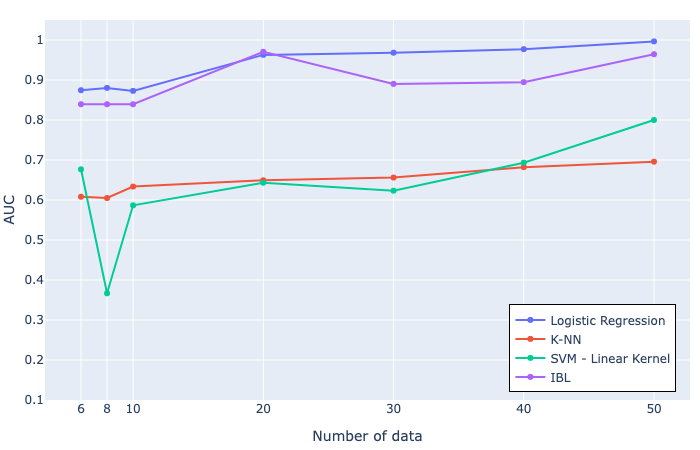}
        \label{fig:t3}
      }
      \caption{Comparison of AUC for each training data size on the Titanic dataset for IBL and each machine learning algorithm}
      \label{fig:iblvsml}
    \end{figure}

\begin{figure}[H]
      \centering
      \subfigure[seed=3655]{
        \includegraphics[width=0.3\linewidth]{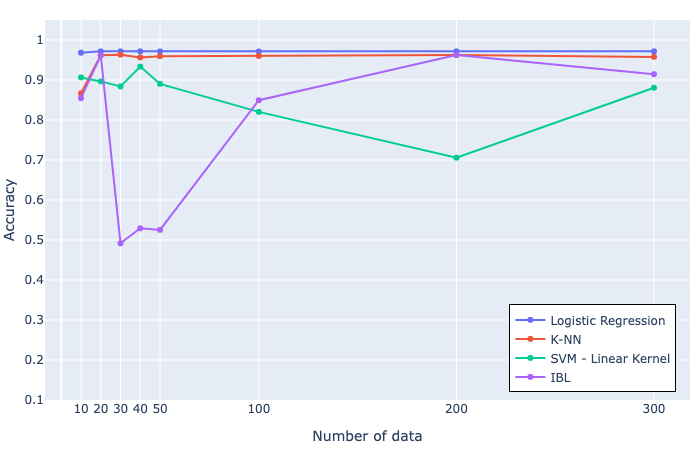}
        \label{fig:p1}
      }
      \subfigure[seed=3656]{
        \includegraphics[width=0.3\linewidth]{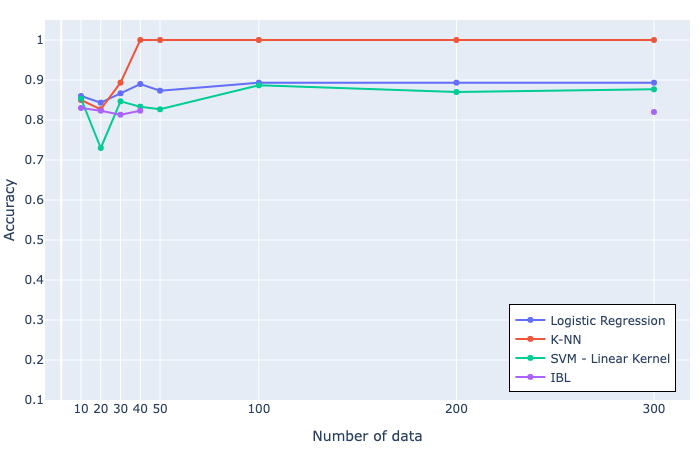}
        \label{fig:p2}
      }
      \subfigure[seed=3657]{
        \includegraphics[width=0.3\linewidth]{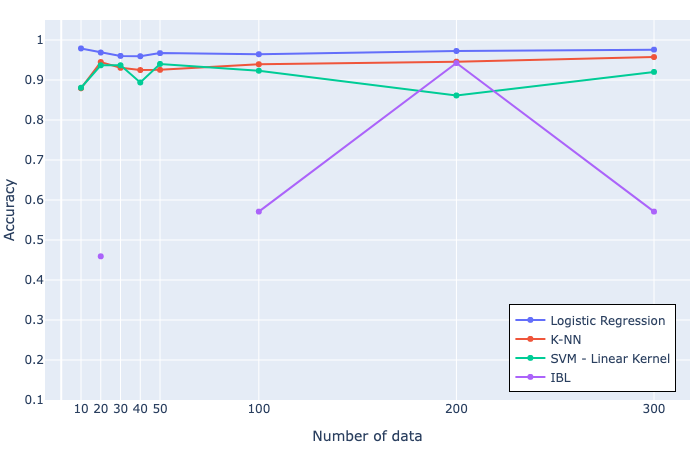}
        \label{fig:p3}
      }
      \caption{CComparison of AUC for each training data size on the pseudo dataset for IBL and each machine learning algorithm}
      \label{fig:iblvsml2}
    \end{figure}

\begin{figure}[H]
      \centering
      \subfigure[seed=3655]{
        \includegraphics[width=0.3\linewidth]{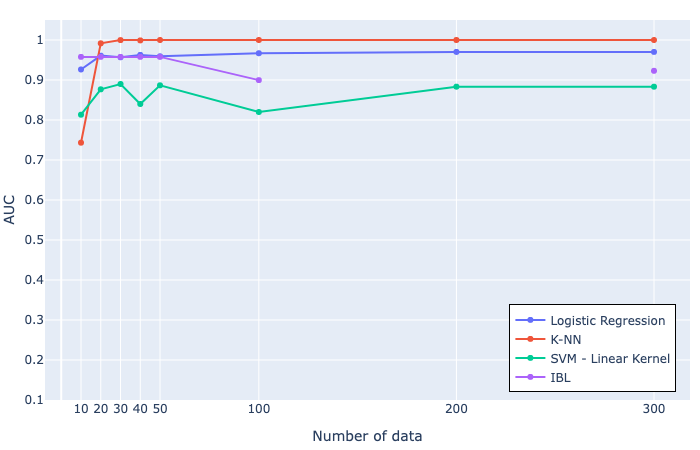}
        \label{fig:m1}
      }
      \subfigure[seed=3656]{
        \includegraphics[width=0.3\linewidth]{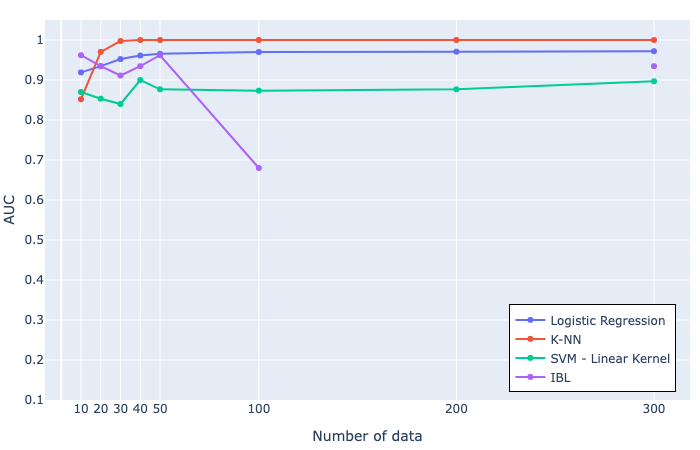}
        \label{fig:m2}
      }
      \subfigure[seed=3657]{
        \includegraphics[width=0.3\linewidth]{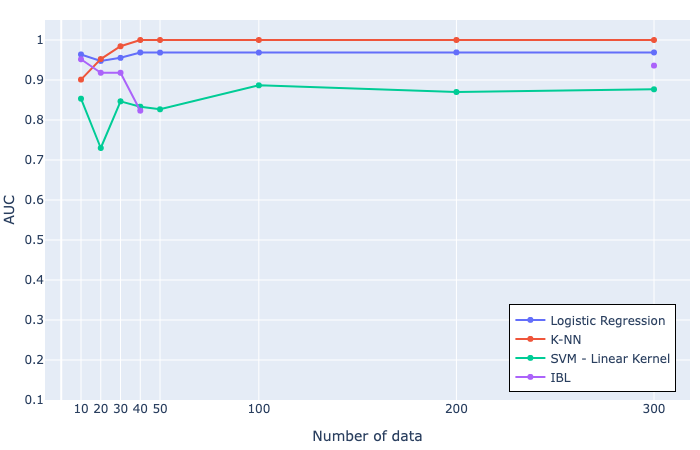}
        \label{fig:m3}
      }
      \caption{Comparison of AUC for each training data size on the moon dataset for IBL and each machine learning algorithm}
      \label{fig:iblvsml3}
    \end{figure}

In the first validation, we compared the AUC of the Code Models generated by IBL with typical machine learning models using the Titanic dataset, the pseudo dataset, and the moon dataset, with various training data size.
 The Logistic Regression, the K-Nearest Neighbors(K-NN), and the Linear Kernel SVM were used as machine learning models for comparison.
In IBL, code models were generated 30 times for each seed value, and the models with the highest AUC were selected for the results. 
If there are NO Code Models that can be successfully generated, the results are not plotted in the graph. 
The following are the validation results for each dataset(Figure \ref{fig:iblvsml}, \ref{fig:iblvsml2}, \ref{fig:iblvsml3}).

\subsubsection*{Titanic dataset}
        
For all seed values and all training data size,  IBL achieves higher AUC than the K-NN and the SVM.
IBL also achieves higher AUC than all machine learning models in some conditions.

\subsubsection*{Pseudo dataset}

In some cases, the pseudo dataset outputs AUC  comparable to other machine learning models, but the overall accuracy is often low. 
Furthermore, depending on the dataset size, there may be cases where viable models are not generated.

\subsubsection*{Moon dataset}

The AUC in the moon dataset often exceed those of the SVM, although they are inferior to those of the K-NN and the logistic regression.

IBL achieves very high AUC and accuracy on the Titanic dataset.
This may be due to the fact that the gpt-4-0613 model includes codes that performed EDA on the Kaggle dataset in the training data.
It is also possible that the model predicts Survived, the objective variable of the Titanic dataset, based on the name of each explanatory variable alone. 
Conversely, it is impossible to predict the value of the objective variable from the names of the explanatory variables for the pseudo dataset and the moon dataset, and it is also considered very difficult to identify that the data are generated using libraries such as sklearn in a very small dataset size. It is therefore considered that the LLM outputs logic for inference based only on the features from the data.

Additionally, the high accuracy with the Titanic dataset might be attributed to the fact that it contains more categorical variables compared to the pseudo datasets or the moon dataset. 
This could have made it easier for the LLM to generate the logic for prediction when creating Code Models. 
To determine whether continuous or discrete explanatory variables are more easily handled by IBL, further verification with different datasets will be necessary.

Some of the generated Code Models are attached in the Appendices.

\subsection{IBL vs ICL}
\begin{figure}[H]
  \centering
  \includegraphics[width=0.7\textwidth]{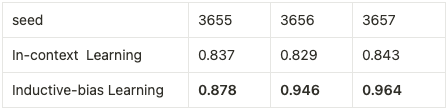}
  \caption{Comparison of AUC between ICL and IBL in Titanic dataset}
  \label{fig:iblvsicl}
\end{figure}

The second validation compares ICL with IBL. 
The data for the comparison was Kaggle's Titanic dataset. 
These training data were sampled with three different seed values from the Titanic dataset, while the other data was used as test data. Below are the validation results(Figure \ref{fig:iblvsicl}).

Figure \ref{fig:iblvsicl} shows that IBL can be predicted with an AUC higher than ICL for all seed values. 
Furthermore, IBL can predict with an AUC that is more than 10 points higher than ICL, depending on the seed value. 
These results suggest that ICL may be making predictions based on training data input to the prompt, but when dataset size and features are large, it is difficult to find the characteristics of the entire dataset, and some data may be used to make predictions similar to some kind of over-trained model.
On the other hand, IBL uses all of the training data for code generation, so it may be able to achieve higher accuracy than ICL due to its higher generalization performance.

Moreover, the contents of the Titanic dataset and the code used to analyze it are likely to have been used in the training of gpt-4-0613, which may have affected the code generation in IBL and the prediction in ICL. It is an interesting topic to see how much IBL and ICL are affected by the LLM training data on which they are based.

In this verification, we were unable to run the same test data multiple times on the same training data, or use other datasets or seed values, because the cost of the OpenAI API is very high when running ICL. We would like to work on these validations in the future.

\section{Future works}
IBL has the potential to be a paradigm shift for building new ``machine learning algorithm''. 
The following are summaries of validation items that are expected to become important in the future.

\subsubsection*{Improved accuracy of IBL}
The Code Model with accuracy that exceeds that of ICL, when compared to machine learning models, accuracy is often inferior as the dataset size increases.
This is because machine learning models generally improve in accuracy as the dataset size increases, but we have not observed such a phenomenon with IBL. 
This may be due to the fact that the training data input to the LLM are not sufficiently analyzed and the necessary patterns for prediction are not output, as in CoT(Chain-of-Thought)\cite{wei2022chain} , where the first instruction is to output text-based elements for prediction from the training data, and the second instruction is to output unknown data based on these texts. 
Then, a Code Model is created to predict unknown data based on the text, which is expected to improve the prediction accuracy.

Additionally, it is expected that more accurate models can be generated by incorporating statistical learning theory techniques like Bagging and Boosting, re-editing the created Code Model, and optimizing the prompts\cite{sordoni2023deep} that are input together with the training data.

We are also considering the possibility of IBL replacing socially implemented machine learning algorithms, such as the LightGBM\cite{ke2017lightgbm}. Current Code Models generated by IBL have problems such as inferior accuracy and unstable output of Code Models.
Conversely, if these problems can be solved to some extent, the performance of the Code Model will be useful enough for social implementation in terms of model inference speed and model interpretability.

\subsubsection*{Understanding of IBL capabilities}
It remains to be analyzed how IBL infers a logic for inference from the training data to create Code Models and how good the inference logic of the created Code models is.
For the latter, the inverstigate by \cite{garg2022can} suggests that ICL can approximate linear functions with accuracy comparable to that of least squares estimators. 
This may allow us to verify how accurately LLMs can obtain regression coefficients from data using IBL.

\subsubsection*{Analysis of the generated Code model}
More detailed analysis of the generated Code Models are needed, as they are still unclear what kind of logics are used inside the Code models and what kind of relationships exists between the number of data prompts (IBL training data size) and an internal logic. 
The former question is about the types of Code Models that can be generated, and while logistic regression-like structures and conditional branching with if statements were generated in most cases in this verification, it would be very interesting to see what other types of structures can be generated (e.g., nonlinear relationships involving exponential functions). 
It would be very interesting to see what other structures can be generated (e.g., nonlinear relationships involving exponential functions, etc.). The relationship between the latter and how the Code Model changes as the number of input data is increased is not yet known.

\section{Conclusion}
In this paper, we proposed a new learning method called ``Inductive-bias Learning ''(IBL), which uses LLM to create inference models. 
We found that the inference models generated by this learning method, Code Models, sometimes outperform conventional machine learning models. 
We also show that the accuracy of IBL exceeds that of ICL on the Titanic datasets. We believe that the predictive models generated by IBL (Code Model) have the potential to replace existing machine learning models in terms of interpretability and inference speed as their accuracy improves in the future. 
Furthermore, we believe that IBL may lead to a better understanding for ICL, which still has many unknowns.

\bibliography{./ibl.bib}
\bibliographystyle{IEEEtranS}

\newpage

\appendix
\section{Appendices}
The following is an example of the execution of the Code Model generated in this experiment.

\subsection{Successful execution Code Model}
Some examples of successfully created Code Models that can predict with high AUC for each data set are shown below.

\subsubsection*{Titanic dataset}       
seed=3656, Number of training data 50, AUC=0.9455
\begin{lstlisting}[language=Python, caption=$\mathrm{seed=3656, \ Number \ of \ training \ data \ 50}$]
import numpy as np
def predict(x):
    df = x.copy()
    output = []
    for index, row in df.iterrows():
        # Do not change the code before this point.
        # Please describe the process required to make the prediction below.

        # The logic here is to give more weightage to the features that are more likely to result in survival.
        # For example, passengers in first class (pclass=1), female passengers (sex_female=1), passengers who embarked at Cherbourg (embarked_C=1), 
        # passengers who are alone (alone_True=1), passengers who are not adult males (adult_male_False=1), passengers who are women (who_woman=1), 
        # passengers in Deck B, C, D, E (deck_B=1, deck_C=1, deck_D=1, deck_E=1) and passengers who embarked at Cherbourg (embark_town_Cherbourg=1) 
        # are more likely to survive. Therefore, these features are given more weightage in the prediction.
        y = row['pclass']*(-0.15) + row['age']*(-0.02) + row['sibsp']*(-0.05) + row['parch']*0.05 + row['fare']*0.002 + row['sex_female']*0.3 + row['sex_male']*(-0.3) + row['embarked_C']*0.1 + row['embarked_Q']*0.05 + row['embarked_S']*(-0.05) + row['alive_no']*(-0.3) + row['alive_yes']*0.3 + row['alone_False']*(-0.1) + row['alone_True']*0.1 + row['adult_male_False']*0.15 + row['adult_male_True']*(-0.15) + row['who_child']*0.1 + row['who_man']*(-0.15) + row['who_woman']*0.15 + row['class_First']*0.15 + row['class_Second']*0.05 + row['class_Third']*(-0.1) + row['deck_A']*0.05 + row['deck_B']*0.1 + row['deck_C']*0.1 + row['deck_D']*0.1 + row['deck_E']*0.1 + row['deck_F']*0.05 + row['deck_G']*0.05 + row['embark_town_Cherbourg']*0.1 + row['embark_town_Queenstown']*0.05 + row['embark_town_Southampton']*(-0.05)

        # The output is then passed through a sigmoid function to convert it into a probability between 0 and 1.
        y = 1 / (1 + np.exp(-y))

        # Do not change the code after this point.
        output.append(y)
    return np.array(output)
\end{lstlisting}
seed=3657, Number of training data 20, AUC = 0.9705
\begin{lstlisting}[language=Python, caption=$\mathrm{seed=3657, \ Number \ of \ training \ data \ 20}$]
import numpy as np

def predict(x):
    df = x.copy()
    output = []
    for index, row in df.iterrows():
        # Do not change the code before this point.
        # Please describe the process required to make the prediction below.

        # The logic here is to give higher weightage to the features that are more likely to result in survival (target=1)
        # and lower weightage to the features that are more likely to result in non-survival (target=0).
        # The weights are determined based on the data provided.

        y = 0.0
        y += row['pclass'] * -0.15
        y += row['age'] * -0.02
        y += row['sibsp'] * -0.05
        y += row['parch'] * 0.05
        y += row['fare'] * 0.002
        y += row['sex_female'] * 0.3
        y += row['sex_male'] * -0.3
        y += row['embarked_C'] * 0.1
        y += row['embarked_Q'] * 0.05
        y += row['embarked_S'] * -0.05
        y += row['alive_no'] * -0.5
        y += row['alive_yes'] * 0.5
        y += row['alone_False'] * 0.1
        y += row['alone_True'] * -0.1
        y += row['adult_male_False'] * 0.2
        y += row['adult_male_True'] * -0.2
        y += row['who_child'] * 0.2
        y += row['who_man'] * -0.2
        y += row['who_woman'] * 0.2
        y += row['class_First'] * 0.2
        y += row['class_Second'] * 0.1
        y += row['class_Third'] * -0.1
        y += row['deck_A'] * 0.05
        y += row['deck_B'] * 0.1
        y += row['deck_C'] * 0.1
        y += row['deck_D'] * 0.1
        y += row['deck_E'] * 0.1
        y += row['deck_F'] * 0.05
        y += row['deck_G'] * 0.05
        y += row['embark_town_Cherbourg'] * 0.1
        y += row['embark_town_Queenstown'] * 0.05
        y += row['embark_town_Southampton'] * -0.05

        # Convert the final score to a probability using the sigmoid function
        y = 1 / (1 + np.exp(-y))

        # Do not change the code after this point.
        output.append(y)
    return np.array(output)
\end{lstlisting}

\subsubsection*{Pseudo dataset}  
 seed=3655, Number of training data 10, AUC=0.8546
\begin{lstlisting}[language=Python, caption=$\mathrm{seed=3655, \ Number \ of \ training \ data \ 10}$]
import numpy as np
import pandas as pd

def predict(x):
    df = x.copy()
    output = []
    for index, row in df.iterrows():
        # Do not change the code before this point.
        # Please describe the process required to make the prediction below.

        # Calculate the sum of the absolute values of the features
        sum_abs = abs(row['a']) + abs(row['b']) + abs(row['c']) + abs(row['d'])

        # Calculate the sum of the positive values of the features
        sum_pos = max(0, row['a']) + max(0, row['b']) + max(0, row['c']) + max(0, row['d'])

        # Calculate the probability as the ratio of the sum of positive values to the sum of absolute values
        y = sum_pos / sum_abs if sum_abs != 0 else 0.5

        # Do not change the code after this point.
        output.append(y)
    return np.array(output)
\end{lstlisting}
seed=3655, Number of training data 300, AUC=0.9143
\begin{lstlisting}[language=Python, caption=$\mathrm{seed=3655, \ Number \ of \ training \ data \ 300}$]
import numpy as np

def predict(x):
    df = x.copy()
    output = []
    for index, row in df.iterrows():
        # Do not change the code before this point.
        # Please describe the process required to make the prediction below.
        # Here we are using a simple heuristic based on the observation that higher values of 'a' and 'c' and lower values of 'b' and 'd' tend to correspond to a target of 1.
        # This is a very simplistic approach and would likely be improved with a more sophisticated model.
        y = 0.5 + 0.5 * (row['a'] - row['b'] + row['c'] - row['d']) / 4
        y = max(min(y, 1), 0)  # Ensure y is between 0 and 1

        # Do not change the code after this point.
        output.append(y)
    return np.array(output)
\end{lstlisting}
                
\subsubsection*{Moon data}
seed=3655, Number of training data 30, AUC=0.9148
\begin{lstlisting}[language=Python, caption=$\mathrm{seed=3655, \ Number \ of \ training \ data \ 30}$]
import numpy as np
import pandas as pd

def predict(x):
    df = x.copy()
    output = []
    for index, row in df.iterrows():
        # Do not change the code before this point.
        # Please describe the process required to make the prediction below.

        # Here we are using a simple linear regression model to predict the target.
        # The coefficients are chosen based on the observation of the data.
        # The intercept is set to 0.5 to make sure the output is between 0 and 1.
        y = 0.5 + 0.3 * row['Feature_1'] - 0.2 * row['Feature_2']

        # Make sure the output is between 0 and 1.
        y = max(0, min(1, y))

        # Do not change the code after this point.
        output.append(y)
    return np.array(output)
\end{lstlisting}
seed=3657, Number of training data 10, AUC=0.9518
\begin{lstlisting}[language=Python, caption=$\mathrm{seed=3657, \ Number \ of \ training \ data \ 10}$]
import numpy as np
import pandas as pd

def predict(x):
    df = x.copy()
    output = []
    for index, row in df.iterrows():
        # Do not change the code before this point.
        # Please describe the process required to make the prediction below.

        # Here we are using a simple linear regression model for prediction.
        # The coefficients are calculated based on the given data.
        # The intercept is assumed to be 0 for simplicity.
        # The coefficients for Feature_1 and Feature_2 are 0.4 and -0.6 respectively.
        # These coefficients are chosen such that the predicted value is close to 1 for target=1 and close to 0 for target=0.
        y = 0.4 * row['Feature_1'] - 0.6 * row['Feature_2']

        # The predicted value is then converted to a probability using the sigmoid function.
        y = 1 / (1 + np.exp(-y))

        # Do not change the code after this point.
        output.append(y)
    return np.array(output)
\end{lstlisting}
          
\subsection{Failed execution Code Model}
Code Models that fail to execute could be roughly divided into the following three types of patterns:
\begin{enumerate}
  \item Failure to import generated Code Model
  \item If the import succeeds in executing but the predict method fails
  \item The predict method is successfully executed, but the predicted probability value is incorrect
\end{enumerate}
Below are examples of Code Models that failed to execute in the verification of pseudo datasets.

First is the first pattern. In the following, the Python code itself is generated, but the entire code block is enclosed in backquotes, or a natural sentence describing the generated Python code is included at the beginning of the sentence, causing a syntax error and resulting in import failure.

\begin{lstlisting}[language=Python, caption=Code Model of pattern 1 that failed to execute ]
Here is a simple Python code that uses a basic logistic regression model to predict the probability that the "target" of the unknown data is 1. This code does not use any existing machine learning model, but rather implements the logistic regression model from scratch.

```python
import numpy as np
import pandas as pd

def sigmoid(x):
    return 1 / (1 + np.exp(-x))

def predict(x):
    df = x.copy()
    output = []
    for index, row in df.iterrows():
        # Do not change the code before this point.
        # Please describe the process required to make the prediction below.
        
        # Define the logistic regression model
        z = 0.1*row['a'] + 0.2*row['b'] + 0.3*row['c'] + 0.4*row['d']
        y = sigmoid(z)
        
        # Do not change the code after this point.
        output.append(y)
    return np.array(output)
```

In this code, the `sigmoid` function is used to map any real-valued number into the range [0, 1], which can be interpreted as probabilities. The `predict` function applies the logistic regression model to each row of the input DataFrame `x` to predict the probability that the "target" of the unknown data is 1. The coefficients of the logistic regression model (0.1, 0.2, 0.3, 0.4) are chosen arbitrarily and should be adjusted based on the actual data to improve the prediction accuracy.
\end{lstlisting}

The second pattern is an implementation that is fine as Python code, but causes a behavior error when the data is actually given and processed.

In the following the Code Models process, there are a part where predictions are converted to probability values. 
In the denominator of this part, although y is a scalar value, min and max are calculated and the difference is taken, resulting in a division by zero and all the values become nan.

\begin{lstlisting}[language=Python, caption=Code Model of pattern 2 that failed to execute]
import numpy as np
import pandas as pd

def predict(x):
    df = x.copy()
    output = []
    for index, row in df.iterrows():
        # Do not change the code before this point.
        # Please describe the process required to make the prediction below.
        
        # Calculate the sum of the absolute values of the first four columns
        sum_abs = np.sum(np.abs(row[['a', 'b', 'c', 'd']]))
        
        # Calculate the mean of the first four columns
        mean = np.mean(row[['a', 'b', 'c', 'd']])
        
        # Calculate the standard deviation of the first four columns
        std = np.std(row[['a', 'b', 'c', 'd']])
        
        # Calculate the probability using the formula: (sum_abs + mean) / (std + 1)
        # The "+1" in the denominator is to prevent division by zero
        y = (sum_abs + mean) / (std + 1)
        
        # Normalize the probability to be between 0 and 1
        y = (y - np.min(y)) / (np.max(y) - np.min(y))
        
        # Do not change the code after this point.
        output.append(y)
    return np.array(output)
\end{lstlisting}

The third pattern is where the predictions are not properly normalized and exceed 1, or the predictions are negative.

In the following example, a min/max calculation should be performed for y in the normalization calculation section, but a min/max calculation is performed for df.

\begin{lstlisting}[language=Python, caption=Code Model of pattern 3 that failed to execute]
import numpy as np
import pandas as pd


def predict(x):
    df = x.copy()
    output = []
    for index, row in df.iterrows():
        # Do not change the code before this point.
        # Please describe the process required to make the prediction below.

        # Calculate the sum of the absolute values of the first four columns
        sum_abs = abs(row["a"]) + abs(row["b"]) + abs(row["c"]) + abs(row["d"])

        # Calculate the sum of the squares of the first four columns
        sum_squares = row["a"] ** 2 + row["b"] ** 2 + row["c"] ** 2 + row["d"] ** 2

        # Calculate the product of the first four columns
        product = row["a"] * row["b"] * row["c"] * row["d"]

        # Calculate the average of the first four columns
        average = (row["a"] + row["b"] + row["c"] + row["d"]) / 4

        # Calculate the prediction as a weighted sum of the above four values
        y = 0.2 * sum_abs + 0.3 * sum_squares + 0.1 * product + 0.4 * average

        # Normalize the prediction to the range [0, 1]
        y = (y - df.min()) / (df.max() - df.min())

        # Do not change the code after this point.
        output.append(y)
    return np.array(output)
\end{lstlisting}

\end{document}